\documentclass{article} 
\PassOptionsToPackage{sort&compress}{natbib}
\usepackage{iclr2017_conference,times}
\usepackage{hyperref}
\usepackage{url}
\newcommand{\ignore}[1]{}
\usepackage{bbm}
\usepackage{graphicx}
\usepackage{authblk}
\title{A Compositional Object-Based Approach to Learning Physical Dynamics}

\author[*]{\textbf{Michael B. Chang}}
\author[**]{\textbf{Tomer Ullman}}
\author[*]{\textbf{Antonio Torralba}}
\author[**]{\textbf{Joshua B. Tenenbaum}}
\affil[*]{Department of Electrical Engineering and Computer Science, MIT}
\affil[**]{Department of Brain and Cognitive Sciences, MIT}
\affil[ ]{\texttt {\{mbchang,tomeru,torralba,jbt\}@mit.edu}}

\iclrfinalcopy 

\begin{document}
\maketitle

\begin{abstract}
We present the Neural Physics Engine (NPE), a framework for learning simulators of intuitive physics that naturally generalize across variable object count and different scene configurations. We propose a factorization of a physical scene into composable object-based representations and a neural network architecture whose compositional structure factorizes object dynamics into pairwise interactions. Like a symbolic physics engine, the NPE is endowed with generic notions of objects and their interactions; realized as a neural network, it can be trained via stochastic gradient descent to adapt to specific object properties and dynamics of different worlds. We evaluate the efficacy of our approach on simple rigid body dynamics in two-dimensional worlds. By comparing to less structured architectures, we show that the NPE's compositional representation of the structure in physical interactions improves its ability to predict movement, generalize across variable object count and different scene configurations, and infer latent properties of objects such as mass.
\end{abstract}
\section{Introduction} \label{section:intro}
Endowing an agent with a program for physical reasoning constrains the agent's representation of the environment by establishing a prior on the environment's physics. The agent can leverage these constraints to rapidly learn new tasks, to flexibly adapt to changes in inputs and goals, and to naturally generalize reasoning to novel scenes \citep{lake2016building}. 

For example, a foundational sense of intuitive physics is a prior that guides humans to decompose a scene into objects and carry expectations of object boundaries and motion across different scenarios \citep{spelke1990principles}. Humans perceive balls on a billiard table not as meaningless patches of color but rather as impermeable objects. They expect balls moving toward each other to bounce a certain way after a collision rather than pass through each other, crumble into pieces, or disperse into smoke. Replace one billiard ball with a bowling ball and expectations for ball-to-ball interactions will differ, but the underlying sense of inertia and collisions remain. Arrange immovable wooden obstacles on the table and expectations for how a ball's surface interacts with wood remain constant regardless of how the obstacles are arranged. The ability to plan trajectories in this space without having to relearn physics from scratch each time, regardless of whether there are three balls or eight balls, whether there are obstacles or not, whether obstacles are arranged in one way or another, whether or not the configuration of objects has been seen before, suggests that humans leverage a prior on physics to reason at a level of abstraction where objects, relations, and events are primitive.

This paper explores the question of building this prior into an agent as a program. We view this program as a simulator that takes input provided by a physical scene and the past states of objects, and outputs the future states and physical properties of relevant objects \citep{anderson1990cognitive,goodmanprobmod,battaglia2013simulation}. Our goal is to design a program that naturally generalizes across variable object count and different scene configurations without additional retraining. Our proposed framework, the Neural Physics Engine (NPE), outlines several ingredients useful for realizing these two generalization capabilities. We describe these ingredients in the context of a specific instantiation of the NPE applied to two-dimensional worlds of balls and obstacles.

\subsection{A hybrid design}\label{intro:design}
Two general approaches have emerged in the search for a program that captures common-sense physical reasoning. The top-down approach \citep{battaglia2013simulation,bates2015humans,ullman2014learning,hamrick2011internal,wu2015galileo} formulates the problem as inference over the parameters of a symbolic physics engine, while the bottom-up approach \citep{agrawal2016learning,fragkiadaki2015learning,lerer2016,li2016fall,mottaghi2015newtonian,mottaghi2016happens,sutskever2009recurrent} learns to directly map observations to motion prediction or physical judgments. A program under the top-down approach can generalize across any scenario supported by the entities and operators in its description language. However, it may be brittle under scenarios not supported by its description language, and adapting to these new scenarios requires modifying the code or generating new code for the physics engine itself. In contrast, gradient-based bottom-up approaches can apply the same model architecture and learning algorithm to specific scenarios without requiring the physical dynamics of the scenario to be pre-specified. This often comes at the cost of reduced generality: transferring knowledge to new scenes may require extensive retraining, even in cases that seem trivial to human reasoning. 

The NPE takes a step toward bridging the gap between expressivity and adaptability by combining the strengths of both approaches. The NPE framework is realized as a differentiable physics simulator that combines rough symbolic structure with gradient-based learning. It exhibits several strong inductive biases that are explicitly present in symbolic physics engines, such as a notion of objects-specific properties and object interactions. Implemented as a neural network, the NPE can also flexibly tailor itself to specific object properties and dynamics of a given world through training. By design, it can extrapolate to a variable number of objects and different scene configurations with only spatially and
temporally local computation. 

\subsection{Ingredients useful for generalization}\label{intro:ingredients}
Our framework proposes four key ingredients useful for generalization across variable object count and different scene configurations without additional retraining. The first ingredient is the view of objects as primitives of physical reasoning. The second is a mechanism for selecting context objects given a particular object. Together, these ingredients reflect two natural assumptions about a physical environment: There exist objects and these objects interact in a factorized manner.  

The third and fourth ingredients are factorization and compositionality, which are both applied on two levels: the scene and the network architecture. On the level of the physical scene, the NPE \textit{factorizes} the scene into object-based representations, and \textit{composes} smaller building blocks to form larger objects. This method of representation adapts to scene configurations of variable complexity and shape. On the level of the network architecture, the NPE explicitly reflects a causal structure in object interactions by \textit{factorizing} object dynamics into pairwise interactions. The NPE models the future state of a single object as a function \textit{composition} of the pairwise interactions between itself and other context objects in the scene. This structure serves to guide learning towards object-based reasoning and is designed for physical knowledge to transfer across variable number objects anywhere in the scene.

\subsection{A step towards emulating a general-purpose physics engine}\label{intro:step}
While previous bottom-up approaches (Sec. \ref{section:related_work}) have coupled learning vision and learning physical dynamics, we take a different approach for two reasons. First, we see that disentangling the visual properties of an object from its physical dynamics is a step toward achieving the generality of a physics engine. Both vision and dynamics are necessary, but we believe that keeping these functionalities separate is important for common-sense generalization that is robust to cases where the visual appearance changes but the dynamics remain the same. Second, we are optimistic that those two components indeed can be decoupled, that a vision model can map visual input to an intermediate state space, and a dynamics model can evolve objects in that state space through time. For example, there is work in object detection and localization (e.g. \citealt{eslami2016attend}) for extracting position and velocity, as well as work for extracting latent object properties \citep{wu2015galileo,phys101}. Therefore this paper focuses on learning dynamics in that state space, taking a small step toward emulating a general-purpose physics engine, with the eventual goal of building a system that exhibits the compositionality, modularity, and generality of a physics engine whose internal components can be learned through observation.

In Sec. \ref{section:approach} we present a specific instantiation of the NPE that uses a neighborhood mask to select context objects. In Sec. \ref{section:evaluation} we apply that instantiation to investigate variations on two-dimensional worlds of balls and obstacles from the matter-js physics engine \citep{brummittmatter} as a testbed for exploring the NPE's capabilities to model simple rigid-body dynamics. While these worlds are generated from a simplified physics engine, we believe that learning to model such simple physics under the NPE's framework is a first and necessary step towards emulating the full capacity of a general physics engine, while maintaining a differentiability that can allow it to eventually learn complex real-world physical phenomena that would be challenging to engineer into conventional physics engines. This paper establishes that important step.
\section{Approach} \label{section:approach}
\subsection{Neural Physics Engine} \label{section:npe}
\begin{figure*}[t]
  \centering
 \includegraphics[width=0.7\textwidth,height=\textheight,keepaspectratio]{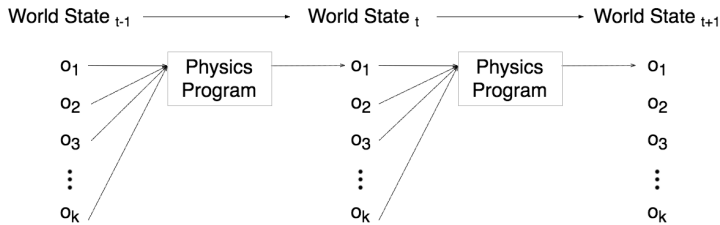}
  \caption{\textbf{Physics Programs:} We consider the space of physics programs over object-based representations under physical laws that are Markovian and translation-invariant. We consider each object in turn and predict its future state conditioned on the past states of itself and its context objects.}
  \label{figure:overview}
\end{figure*}

We consider in detail a specific instantiation of the NPE that uses a neighborhood mask to select context objects. This section discusses each of the four ingredients of the NPE framework, that, when combined, comprise a neural network-based physics simulator that learns from observation.
\paragraph{Object-Based Representations} \label{section:scenario}
We make two observations (Fig. \ref{figure:overview}) in our factorization of the scene. The first regards spatially local computation. Because physics does not change across inertial frames, it suffices to separately predict the future state of each object conditioned on the past states of itself and the other objects in its neighborhood, similar to \citet{fragkiadaki2015learning}. Sec. \ref{section:walls} shows that when large structures are represented as a composition of smaller objects, a spatially local attention window helps achieve invariance to scene configuration. The second observation regards temporally local computation. Because physics is Markovian, this prediction need only be for the immediate next timestep, which we show in Sec. \ref{section:evaluation} is enough to predict physics effectively over long timescales. Given these two observations, it is natural to choose an object-based state representation. A state vector comprises  extrinsic properties (position, velocity, orientation, angular velocity), intrinsic properties (mass, object type, object size), and global properties (gravitational, frictional, and pairwise forces) at a given time instance. 

\paragraph{Pairwise Factorization} Letting a particular object be the \textit{focus} object $f$ and all other objects in the scene be \textit{context} objects $c$, the NPE models the focus object's velocity $v^{[t+1]}_{f}$ as a composition of the pairwise interactions between itself and other neighboring context objects in the scene during time $t-1$ and $t$. This input is represented as pairs of object state vectors $\{(o_{f}, o_{c_1})^{[t-1,t]}, (o_{f}, o_{c_2})^{[t-1,t]},...\}$. As shown in Fig. \ref{figure:models}b, the NPE composes an encoder function and a decoder function. The encoder function $f_{enc}$ summarizes the interaction of a single object pair. The sum of encodings of all pairs is then concatenated with the focus object's past state as input to the decoder function. The focus object is a necessary input to the decoder because if there are no neighboring context objects, the summed encoder output would be zero. The decoder function then predicts the focus object's velocity $v^{[t+1]}_{f}$. In practice, the NPE predicts the change $\Delta v$ between $t$ and $t+1$ to compute $v^{[t+1]} = v^{[t]} + \Delta v$, and updates position using the velocity as a first-order approximation\footnote{The NPE as currently implemented also predicts angular velocity along with velocity, but for the experiments in this paper we always set angular velocity, as well as gravity, friction, and pairwise forces, to zero. We included these parameters in the implementation because in future work we are planning to test situations and scenarios in which angular velocity is important, such as block towers, magnetism. However, in the current work they are vestigial and set to zero and do not appear in the evaluation.}. We predict velocity rather than position to help avoid memorizing the environment; training the network to predict position conditions the network on the worlds in the training domain, making it more difficult to transfer knowledge across environments. We do not include acceleration in the state representation because position and velocity fully parametrize an object's state. Thus acceleration (e.g. collisions) can be learned by observing velocity for two consecutive timesteps, hence our choice for two input timesteps. We explored longer input durations as well and found no additional benefit.

\paragraph{Context Selection} Each $\left(o_{f}, o_{c}\right)$ pair is selected to be in the set of neighbors of $f$ by the neighborhood masking function $\mathbbm{1}\left[||p_{c}-p_{f})|| < N(o_{f})\right]$, which takes value $1$ if the Euclidean distance between the positions $p_{f}$ and $p_{c}$ of the focus and context object respectively at time $t$ is less the neighborhood threshold $N(o_{f})$. Many physics engines use a collision detection scheme with two phases. \textit{Broad phase} is used for computational efficiency and uses a neighborhood threshold to select objects that might, but not necessarily will, collide an object. \textit{Narrow phase} performs the actual collision detection on that smaller subset of objects and also resolves the collisions for the objects that do collide. Analogously, our neighborhood mask implements broad phase, and the NPE implements narrow phase. The mask only constrains the search space of context objects, and the network figures out how to detect and resolve collisions. This mask is a specific case of a more general attention mechanism to select contextual elements of a scene.

\paragraph{Function Composition} Symbolic physics engines evolve objects through time based on dynamics that dictate their independent behavior (e.g. inertia) and their behavior with other objects (e.g. collisions). Notably, in a particular object's reference frame, the forces it feels from other objects are additive. The NPE architecture incorporates several inductive biases that reflect this recipe. The composition of $f_{enc}$ and $f_{dec}$  induce a causal structure on the pairs of objects. We provide a loose interpretation of the encoder output $e_{c,f}$ as the \textit{effect} of object $c$ on object $f$, and require that these effects are additive as forces are. This design allows the NPE to scale naturally to different numbers of neighboring context objects. These inductive biases have the effect of strongly constraining the space of possible simulators that the NPE can learn, focusing on compositional programs that reflect pairwise causal structure in object interactions.

\subsection{Baselines}
\begin{figure*}[t]
  \centering
 \includegraphics[width=0.8\textwidth,height=\textheight,keepaspectratio]{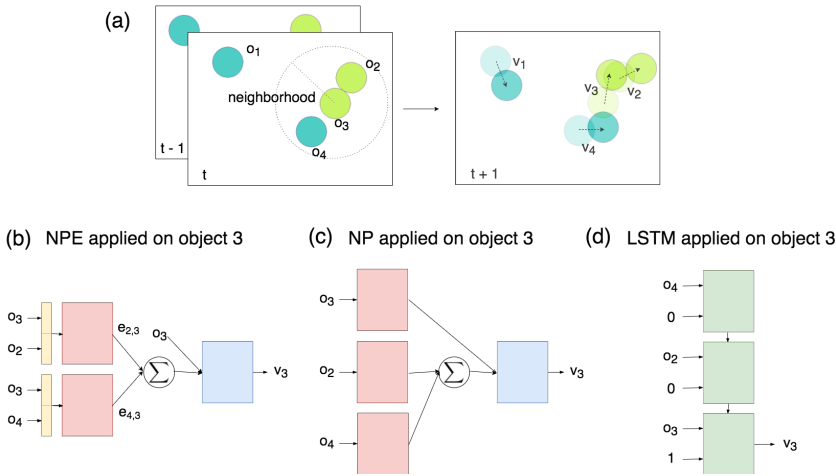}
  \caption{\textbf{Scenario and Models:} This figure compares the NPE, the NP and the LSTM architectures in predicting the velocity of object 3 for an example scenario [\textbf{a}] of two heavy balls (cyan) and two light balls (yellow-green). Objects 2 and 4 are in object 3's neighborhood, so object 1 is ignored. [\textbf{b}]: The NPE encoder consists of a pairwise layer (yellow) and a feedforward network (red) and its decoder (blue) is also a feedforward network. The input to the decoder is the concatenation of the summed pairwise encodings and the input state of object 3. [\textbf{c}]: The NP encoder is the same as the NPE encoder, but without the pairwise layer. The NP decoder is the same as the NPE decoder. The input to the decoder is the concatenation of the summed context encodings and the encoding of object 3. [\textbf{d}]: We shuffle the context objects inputted into the LSTM and use a binary flag to indicate whether an object is a context or focus object.}
  \label{figure:models}
\end{figure*}
The purpose of contrasting the NPE with the following two baselines is to illustrate the benefit of pairwise factorization and function composition, which are the key architectural features of the NPE. As the architectures for both baselines have been shown to work well in similar tasks, it is not immediately clear whether the NPE's assumptions are useful or necessary, so these are good baselines for comparison. Viewed in another way, comparing with these baselines is a lesion study on the NPE because each baseline lacks an aspect of the NPE structure.

\paragraph{No-Pairwise} The No-Pairwise (NP) baseline is summarized by Fig. \ref{figure:models}c. It is very similar to the NPE but does not compute pairwise interactions; otherwise its encoder and decoder are the same as the NPE's. Therefore the NP most directly highlights the value of the NPE's pairwise factorization. The NP is also a Markovian variant of the Social LSTM \citep{alahisocial}; it sums the encodings of context objects after encoding each object independently, similar to the Social LSTM's ``social pooling.'' Information for modeling how objects interact would only be present after the encoding step. A possible mechanism for predicting dynamics with the NP is if the encoder's object encoding consists of an abstract object representation and a force field created by that object. Therefore the decoder could apply the sum of the force fields of all context objects to the focus object's abstract object representation to predict the focus object's velocity. As \citet{alahisocial} has demonstrated the Social LSTM's performance in modeling human trajectories, it would be interesting to see how the same architectural assumptions perform for the physics of moving objects. 

\paragraph{LSTM} Long Short-Term Memory (LSTM) networks \citep{hochreiter1997long} have been shown to sequentially attend to objects \citep{eslami2016attend}, so it is interesting to test whether a LSTM is well-suited for modeling object interactions, when the object states are explicitly given as input. From a cognitive science viewpoint, an LSTM can be interpreted as a serial mechanism in object tracking \citep{pylyshyn2006dynamics}. Our LSTM architecture (Fig. \ref{figure:models}d) accepts the state of each context object until the last step, at which it takes in the focus object's state and predicts its velocity. Because the LSTM moves through the object space sequentially, its lack of factorized compositional structure highlights the value of the NPE's function composition of the independent interactions between an object and its neighbors. Our notion of compositionality treats each object and pairwise interaction as independently encapsulated in a separate computational entity that can be reused and rearranged; the NPE encoder is a function that is applied to each $(o_f, o_c)$ pair. This function encapsulates this computation and can be repeatedly applied to all neighboring context objects equally, such that the NPE composes this repeated encoding function with the decoder function to predict velocity. The LSTM does not exhibit this notion of compositionality because it is not designed to take advantage of the factorized structure of the scene. Unlike the NPE and NP, the LSTM's structure does not differentiate between focus and context object, so we add a flag to the state representation to indicate to whether an object is a context or focus object. We shuffle the order of the context objects to account for an ordering bias.
\section{Experiments} \label{section:evaluation}
Object-based representations (ingredient 1) are necessary for the other three ingredients, and having explained the motivation for object-based representations in Sec. \ref{intro:step} and Sec. \ref{section:npe}, we now analyze the other three ingredients in the context of several experiments. In the prediction task (Sec. \ref{section:prediction}), we first test if the NPE is even capable of predicting physics when the number of objects is held constant. In the generalization task (Sec. \ref{section:generalization}), we test the NPE's capability to generalize across variable object count. In the inference task (Sec. \ref{section:mass}), we test if the NPE can be inverted to infer mass in both the prediction and generalization settings.  In these experiments, we compare against the NPE-NN, a modified NPE without the neighborhood mask, to analyze the context selection mechanism (ingredient 2), the NP to analyze factorization (ingredient 3), the LSTM to analyze compositionality (ingredient 4). Sec. \ref{section:nbrhd_mask} analyzes the neighborhood mask in depth. We test the NPE's capability to generalize across different scene configurations in Sec. \ref{section:walls}.

Using the matter-js physics engine, we evaluate the NPE on worlds of balls and obstacles. These worlds exhibit nonlinear dynamics and support a wide variety of scenarios. Bouncing balls have been of interest in cognitive science to study causality and counterfactual reasoning, as in \cite{gerstenberg2012noisy}. We trained on 3-timestep windows in trajectories of 60 timesteps (10 timesteps $\approx$ 1 second). For a world of $k$ objects, we generate 50,000 such trajectories. For experiments where we train on multiple worlds together, we shuffle the examples across all training worlds and train without a curriculum schedule. All worlds have a vertical dimension of 600 pixels and a horizontal dimension of 800 pixels, and we constrain the maximum velocity of an object to be 60 pixels/second. We normalize positions to $[0,1]$ by dividing by the horizontal dimension, and we normalize velocities to $[-1,1]$ by dividing by the maximum velocity. 

\begin{figure}[t]
  \centering
 \includegraphics[width=\textwidth,height=\textheight,keepaspectratio]{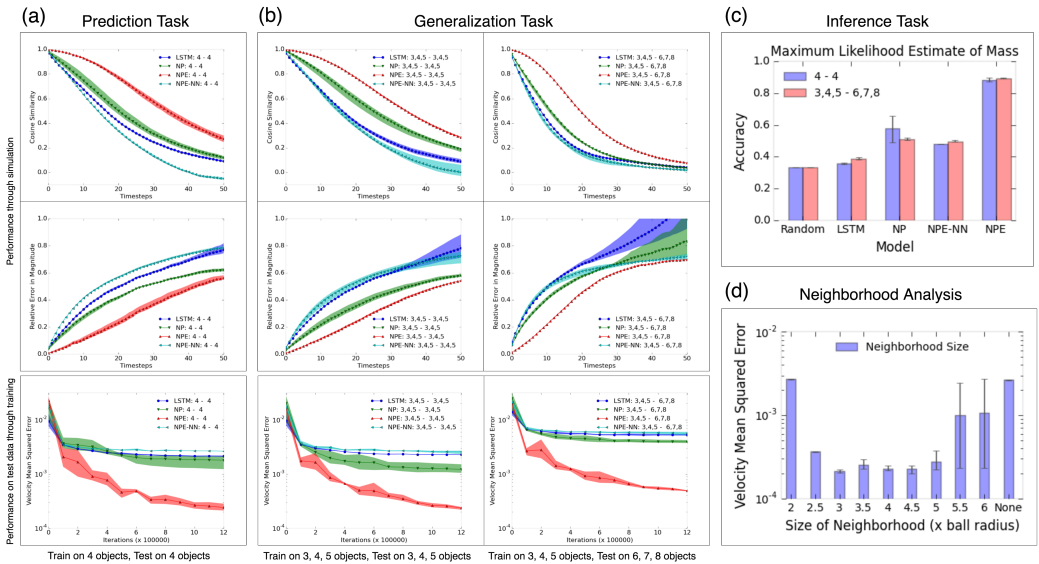}
  \caption{\textbf{Quantitative evaluation (balls):} [\textbf{a,b}]: Prediction and generalization tasks. \textit{Top two rows}: The cosine similarity and the relative error in magnitude. \textit{Bottom row}: The MSE of velocity on the test set over the course of training. Because these worlds are chaotic systems, it is not surprising that all predictions diverge from the ground truth with time, but NPE consistently outperforms the other two baselines on all fronts, especially when testing on 6, 7, and 8 objects in the generalization task. The NPE's performance continues to improve with training while the NPE-NN (an NPE without a neighborhood mask, see Sec. \ref{section:nbrhd_mask}), NP and LSTM quickly plateau. We hypothesize that the NPE's structured factorization of the state space guides it from wasting time exploring suboptimal programs. [\textbf{c}]: The NPE's accuracy is significantly greater than the baseline models' in mass inference. Notably, the NPE achieves similar inference performance whether in the prediction or generalization settings, further showcasing its strong generalization capabilities. The LSTM performs poorest, reaching just above random guessing (33\% accuracy). [\textbf{d}]: We analyze the effectiveness of different neighborhood thresholds for the NPE on the constant-mass prediction task. The neighborhood threshold is quite robust from 3 to 5 ball radii.} 
  \label{figure:ball_quantitative}
\end{figure}

Like those of \cite{battaglia2016interaction} the NPE predictions can be effective over long timescales even when the NPE is only trained to predict the immediate next time step. Randomly selected simulation videos can be found at \url{https://goo.gl/BWYuOF}. Plots show results over three independent runs averaged over held-out test data with different random seeds.  As shown in the graphs in Fig. \ref{figure:ball_quantitative} (top two rows) and Fig. \ref{figure:walls_graph}, both the NP and LSTM's predicted trajectories diverge from the ground truth, but for different reasons, which the videos illuminate. While the NP and LSTM fail to predict plausible physical movement entirely, the NPE's predictions initially adhere closely to the ground truth, then slowly diverge due to the accumulation of subtle errors, just as the human perceptual system also accumulates errors \citep{smith2013sources}. However, the NPE preserves the general intuitive physical dynamics that may roughly be consistent with people's intuitive expectations.

\subsection{Prediction task} \label{section:prediction}
We consider simple worlds of four balls of uniform mass (Fig. \ref{figure:ball_quantitative}a). To measure performance in simulation, we visualize the cosine similarity between the predicted velocity and the ground truth velocity as well as the relative error in magnitude between the predicted velocity and the ground truth velocity over 50 timesteps of simulation. The models take timesteps 1 and 2 as initial input, and then use previous predictions as input to future predictions. To measure progress through training, we also display the Mean Squared Error (MSE) on the normalized velocity.

\subsection{Generalization task}
\label{section:generalization}
We test whether learned knowledge of these simple physics concepts can be transferred and extrapolated to worlds with a number of objects previously unseen (Fig. \ref{figure:ball_quantitative}b). The unseen worlds (6, 7, 8 balls) in the test data are combinatorially more complex and varied than the observed worlds (3, 4, 5 balls) in the training data. All objects have equal mass. During simulation, the NPE's predictions are more consistent, whereas the NP and LSTM's prediction begin to diverge wildly towards the end of 50 timesteps of simulation (Fig. \ref{figure:ball_quantitative}b, middle row). The NPE consistently outperforms the baselines by 0.5 to 1 order of magnitude in velocity prediction (Fig. \ref{figure:ball_quantitative}b, bottom row). 

\subsection{Inference task} \label{section:mass}
We now show that the NPE can infer latent properties such as mass. This proposal is motivated by the experiments in \cite{battaglia2013simulation}, which uses a probabilistic physics simulator to infer various properties of a scene configuration. Whereas the physical rules of their simulator were manually pre-specified, the NPE learns these rules from observation. We train on the same worlds used in both the prediction and generalization tasks, but we uniformly sampled the mass for each ball from the log-spaced set $\{1, 5, 25\}$. We chose to use discrete-valued masses to simplify our qualitative understanding of the model's capacity to infer. For future work we would like to investigate continuously valued masses and evaluate with binary comparisons (e.g. "Which is heavier?").

As summarized by Fig. \ref{figure:ball_quantitative}c and Fig. \ref{figure:visual}a, we select scenarios exhibiting collisions with the focus object, fix the masses of all other objects, and score the NPE's prediction under all possible mass hypotheses for the focus object. The prediction is scored against the ground-truth under the same MSE loss used in training. The hypothesis whose prediction yields the lowest error is the NPE's maximum likelihood estimate of the focus object's mass. Outperforming all baselines, the NPE achieves about 90\% accuracy, meaning it has 90\% probability of inferring the correct mass. 

The NPE predicts outputs given inputs and infers inputs given outputs. Though we adopted a particular parametrization of an object, the NPE is not limited to the semantic meaning of the elements of its input, so we expect other latent object properties can be inferred this way. Because the NPE is differentiable, we expect that it can also infer object properties by backpropagating prediction error to its a randomly sampled input. This would be useful for inferring non-categorical values, such as positions of ``invisible'' objects, whose effects are felt but whose positions are unknown.

\subsection{Neighborhood mask} \label{section:nbrhd_mask}
In Fig. \ref{figure:ball_quantitative}d we vary the NPE's neighborhood threshold $N(o_f)$ and evaluate performance on the constant-mass prediction task. $N(o_f)$ is in units of ball radii, so $N(o_f) = 2$ means that a context object is only detected if it is exactly touching the focus object. Because ball radii are 60 pixels and the maximum velocity is 60 pixels per timestep, the maximum distance two balls can initially be before touching at the next timestep is 4 ball radii. Given that velocities were sampled uniformly, it makes sense that the NPE performs well in and is robust\footnote{The results reported in this paper were with $N(o_f) = 3.5$ ball radii, which we found initially with a coarser search than the results in Fig. \ref{figure:ball_quantitative}d, although any threshold in the range $N(o_f) \in [3, 5]$ performs similarly.} to the range $N(o_f) \in [3, 5]$, but performance drops off with smaller and larger $N(o_f)$. It is important to note that different $N(o_f)$ may work better for different domains and object geometries.

We include analysis in the prediction and generalization tasks on an NPE without the neighborhood mask, the NPE-NN (NN = No Neighborhood). The neighborhood mask gives the NPE about an order of magnitude improvement in velocity prediction loss (Fig. \ref{figure:ball_quantitative}a,b: bottom row and Fig. \ref{figure:summary}). While the NPE loss continues to improve through training, the NPE-NN loss quickly plateaus. It is interesting that the NPE-NN performs no better than both the NP and LSTM in predictive error, but outperforms the LSTM in mass inference. These two observations suggest that computing the interactions the focus object shares with each context object is more effective for inferring a property of the focus object than disregarding these factorized effects. They also suggest that the additional spatial structure from constraining the context space with the neighborhood mask prevents the NPE from naively finding associations with objects that cannot influence the focus object.

In our experiments, the neighborhood mask has the additional practical benefit of reducing computational complexity from $O(k)$ to $O(1)$, where $k$ is the number of objects in the scene, because the number of context-focus object pairs the NPE considers is bounded above by the neighborhood mask at a constant number. Though beyond the scope of this work, to extend the functionality of such context selection mechanism to include worlds that contain forces that act from a distance, future instantiations of the NPE may investigate a more general context selection mechanism that can be learned jointly with the other model parameters.

\subsection{Different Scene Configurations} \label{section:walls}
\begin{figure}[t]
  \centering
 \includegraphics[width=\textwidth,height=\textheight,keepaspectratio]{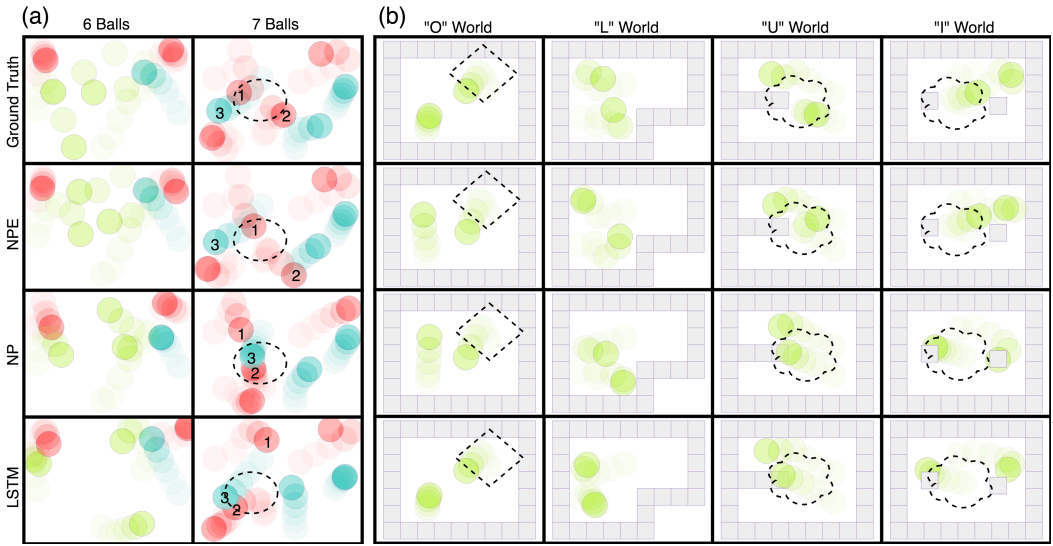}
  \caption{\textbf{Visualizations:} The NPE scales to complex dynamics and world configurations while the NP and LSTM cannot. The masses are visualized as: cyan = 25, red = 5, yellow-green = 1. [\textbf{a}] Consider the collision in the 7 balls world (circled). In the ground truth, the collision happens between balls 1 and 2, and the NPE correctly predicts this. The NP predicts a slower movement for ball 1, so ball 2 overlaps with ball 3. The LSTM predicts a slower movement and incorrect angle off the world boundary, so ball 2 overlaps with ball 3. [\textbf{b}] At first glance, all models seem to handle collisions well in the ``O'' world (diamond), but when there are internal obstacles (cloud), only the NPE can successfully resolve collisions. This suggests that the NPE pairwise factorization handles object interactions well, letting it generalize to different world configurations, whereas the NP and LSTM have only memorized the geometry of the ``O'' world.}
  \label{figure:visual}
\end{figure}
\begin{figure}[t]
  \centering
 \includegraphics[width=\textwidth,height=\textheight,keepaspectratio]{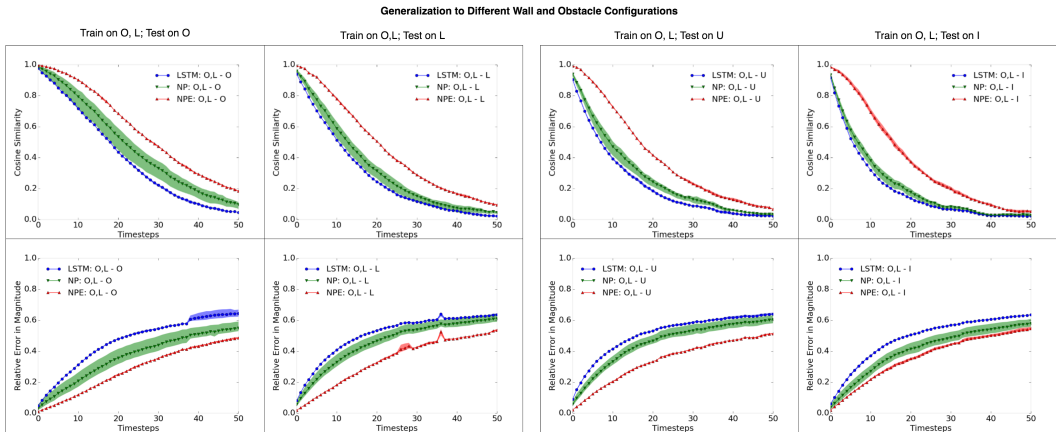}
  \caption{\textbf{Quantitative evalution (walls and obstacles):} The compositional state representation simplifies the physical prediction problem to only be over local arrangements of context balls and obstacles, even when the wall geometries are more complex and varied on a macroscopic scale. Therefore, it is not surprising that the models perform consistently across wall geometries. Note that the NPE consistently outperforms the other models, and this gap in performance increases with more varied internal obstacles for the cosine similarity of the velocity angle. This gap is more prominent in ``L'' and ``U'' geometries for relative error in magnitude.}
  \label{figure:walls_graph}
\end{figure}
We demonstrate representing large structures as a composition of smaller objects as building blocks. This is important for testing the NPE's invariance to scene configuration; the scene configuration should not matter if the underlying physical laws remain the same. These worlds contain 2 balls bouncing around in variations of 4 different wall geometries. ``O'' and ``L'' geometries have no internal obstacles and are in the shape of a rectangle and ``L'' respectively. ``U'' and ``I'' have internal obstacles. Obstacles in ``U'' are linearly attached to the wall like a protrusion, while obstacles in ``I'' have no constraint in position. We randomly vary the position and orientation of the ``L'' concavity and the ``U'' protrusion. We randomly sample the positions of the ``I'' internal obstacles.

We train on conceptually simpler ``O'' and ``L'' worlds and test on more complex ``U'' and ``I'' worlds. Variations in wall geometries adds to the difficulty of this extrapolation task. At most 12 context objects are present in the focus object's neighborhood at a time. The ``U'' geometries have 33 objects in the scene, the most out of all the wall geometries. As shown in Fig. \ref{figure:visual}b and \ref{figure:walls_graph}, the NPE is robust to scenes with internal obstacles, even when it has not observed such scenes during training.

\subsection{Analysis}\label{evaluation:analysis}
We explain the NPE's superior performance in generalization from the perspective of context selection, factorization, and compositionality. By design, all three ingredients transform the testing data distribution to be similar to the training data distribution, such that generalization across variable object count and different scene configurations happens naturally. 

Consider generalizing across variable object count. The neighborhood mask selects context objects such that the NPE need only focus on a bounded subset of the objects regardless of the total number of objects. Factorizing the scene into pairwise interactions induces a causal structure between each context object and the focus object, such that no matter the object count, this causal structure remains consistent because the input is merely a set of object pairs. Composing these pairwise interactions together with a summation encourages the encoder output to be additive, such that the decoder receives the appropriate net effect from the context objects, regardless of how many there are.

Consider generalizing across different scene configurations. Our state representation composes larger structures from smaller objects, just as many real-world objects are composed of smaller components. Therefore, even when wall geometries are complex and varied on a macroscopic scale, the input distribution to the NPE remains roughly the same, because the prediction problem still remains only over objects in a local glimpse the entire scene.
\section{Related Work} \label{section:related_work}
\paragraph{Top-down and bottom-up approaches} A recent set of top-down approaches investigate probabilistic game physics engines as computational models for physical simulation in humans \citep{battaglia2013simulation,bates2015humans,ullman2014learning,hamrick2011internal}. However, these models require a full specification of the physical laws and object geometries. Given such a specification, inferring \textit{how} physical laws compose and apply to a given scenario are their strength, but automatically inferring from visual data \textit{what} physical laws and object properties are present requires more work in inverse graphics \citep{kulkarni2015picture,kulkarni2015deep,whitney2016understanding,kulkarni2014inverse,chen2016infogan} and physics-based visual understanding \citep{wu2015galileo,phys101,brand1997physics}. The NPE builds on top of the key structural assumptions of these top-down approaches, but its differentiable architecture opens a possible path for joint training with a vision model that can automatically adapt to the specific physical properties of the scene.

Bottom-up approaches attempt to bypass the intermediate step of finding physics representations and directly map visual observations to physical judgments \citep{lerer2016,li2016fall,mottaghi2015newtonian,mottaghi2016happens} or passive \citep{lerer2016,srivastava2015unsupervised,sutskever2009recurrent} and action-conditioned \citep{agrawal2016learning,finn2016unsupervised,fragkiadaki2015learning} motion prediction. Because these work historically have not been compositional in nature, they have had limited flexibility to transfer knowledge to conceptually similar worlds where the physics remain the same, but the number of objects or complexity of object configurations varies. Moreover, these approaches above do not infer latent properties as the NPE does.

Other work have taken similar hybrid approaches as the NPE, such as the NeuroAnimator \citep{grzeszczuk1998neuroanimator}, one of the first work to train a neural network to emulate a physics simulator, and the interaction network \citep{battaglia2016interaction}, which learns to simulate physics over a graph of objects and their relations.

\paragraph{Sketching} 
The NPE combines a symbolic structure that assumes generic objects and interactions with a differentiability that allows the specific nature of these interactions to be learned from training. This approach of starting with a general sketch of a program and filling in the specifics is inspired by ideas from the program synthesis community \citep{solar2008program,ellis2015unsupervised,gaunt2016terpret}. Examples of other work that combine symbolic with neural approaches via sketching include graph-based neural networks \citep{jain2016structural,li2015gated,scarselli2009graph} and transforming autoencoders \citep{hinton2011transforming}.

\paragraph{Composing functions for reuse} Just as the NPE repeatedly applies the same encoder to each object pair, iteratively applies itself to each object in the scene as a focus object, and recursively predicts future timesteps using predictions from previous timesteps, employing function reuse to achieve generalization is also featured in work such as \citet{andreas2016learning,lake2015human,reed2015neural,socher2011dynamic,abelson1996structure}. These work all assemble small subprograms to form larger programs. The NPE also dynamically composes its internal modules (encoder and decoder) based on the number of objects and the arrangement of context objects.

\paragraph{Object-based approaches}
\citet{fragkiadaki2015learning} and \citet{battaglia2016interaction} are two notably similar work in the sense that our work and theirs all take an object-based approach to model the bouncing balls environment. Our work was inspired by \citet{fragkiadaki2015learning}'s iterative approach to predicting the motion of each object in turn, conditioned on a context. The key contrast is that their model assumes no relational structure between objects beyond a visual attention window centered around the focus object, whereas ours explicitly processes the interaction between the focus and each context object.

If we compare \textcolor{blue}{\href{https://sites.google.com/site/intuitivephysicsnips15/home}{their simulation videos}} \citep{Intuitiv22:online} to \textcolor{blue}{\href{https://drive.google.com/drive/folders/0BxCJLi4FnT_6QW4tcF94d1doLWs}{ours}}, we see some specific and significant improvements evident in our approach. For example, in their work, the balls appear attracted to each other and to the walls; the balls appear to bounce along the walls even when no attractive force should be present. The balls rarely touch during collisions, but magnetically repel each other when at a short distance. The NPE does not exhibit these behaviors and tends to preserve the intuitive physical dynamics of colliding balls. In addition to these differences, we show strong predictive performance on generalizing to eight balls, five more than the balls in their videos. We also crucially show this performance under stronger generalization conditions, variable mass, and more complex scene configurations.

Recently, \citet{battaglia2016interaction} independently and in parallel developed an architecture that they call the \textit{interaction network} for learning to model physical systems.  They show how such an architecture can apply to several different kinds of physical systems, including n-body gravitational interactions and a string falling under gravity. Like their work, our model can simulate over many timesteps very effectively when only trained for next-timestep prediction, and can generalize to different world configurations and different numbers of objects. 

Compared to the interaction network, a main difference in our architecture is that ours does not take object relations as explicit input, but instead learns the nature of these relations by constraining attention to a neighborhood set of objects. Another difference is in function reuse: we demonstrated that a trained NPE can automatically infer properties of its input such as mass without further retraining. In contrast, they train an additional classifier on top of their model to do inference. Their work also exhibits the four ingredients in our framework, and we view the similarities between their and our work as converging evidence for the utility of object-based representations and compositional model architectures in learning to emulate general-purpose physics engines.

\section{Discussion} \label{section:discussion}
While this paper is not the first to explore learning a physics simulator, here we take the opportunity to highlight the value of this paper's contributions. We hope these contributions can seed further research that builds on the NPE framework this paper proposes. 

We showed that object-based representations, a context selection mechanism, factorization, and compositionality are useful ingredients for learning a physics simulator that generalizes across variable object count and different scene configurations with only spatially and temporally local computation. This generalization is possible because these ingredients transform the testing data distribution to be similar to the training data distribution. 

The NPE makes few but strong assumptions about the nature of objects in a physical environment. These assumptions are inductive biases that not only give the NPE enough structure to help constrain it to model physical phenomena in terms of objects but also are general enough for the NPE to learn physical dynamics almost exclusively from observation.

We applied the NPE to simple two-dimensional worlds of bouncing balls ranging in complexity. We showed that NPE achieves low prediction error, extrapolates learned physical knowledge to previously unseen number of objects and world configurations, and can infer latent properties such as mass. We compared against several baselines designed to test the ingredients of the NPE framework and found superior performance when all these ingredients are combined in the NPE. Though we demonstrated the NPE in the balls environment with nonlinear dynamics and complex scene configurations, the state representation and NPE architecture we propose are quite general-purpose because they assume little about the specific dynamics of a scene. 

This paper works toward emulating a general purpose physics engine under a framework where visual and physical aspects of a scene are disentangled. Next steps include linking the NPE with perceptual models that extract properties such as position and mass from visual input. Learning to simulate is unsupervised learning of the structure of the environment. When a simulator like the NPE is incorporated into an agent in the context of model-based planning and model-based reinforcement learning, it becomes a prior on the environment that guides learning and reasoning. By combining the expressiveness of physics engines and the adaptability of neural networks in a compositional architecture that supports generalization in fundamental aspects of physical reasoning, the Neural Physics Engine is an important step towards lifting an agent's ability to think at a level of abstraction where the concept of physics is primitive.

\subsubsection*{Acknowledgments} We thank Tejas Kulkarni for insightful discussions and guidance. We thank Ilker Yildirim, Erin Reynolds, Feras Saad, Andreas Stuhlm{\"u}ller, Adam Lerer, Chelsea Finn, Jiajun Wu, and the anonymous reviewers for valuable feedback. We thank Liam Brummit, Kevin Kwok, and Guillermo Webster for help with matter-js. This work was supported  MIT's SuperUROP and UROP programs, and by the Center for Minds, Brains and Machines under NSF STC award CCF-1231216 and an ONR grant N00014-16-1-2007.

\bibliography{bibliography}
\bibliographystyle{abbrvnat}

\newpage
\appendix
\section{Implementation}\label{appendix:implementation}
We trained all models using the rmsprop \citep{Tieleman2012} backpropagation algorithm with a Euclidean loss for 1,200,000 iterations with a learning rate of 0.0003 and a learning rate decay of 0.99 every 2,500 training iterations, beginning at iteration 50,000. We used minibatches of size 50 and used a 70-15-15 split for training, validation, and test data.

All models are implemented using the neural network libraries built by \citet{collobert2011torch7,leonard2015rnn}. The NPE encoder consists of a pairwise layer of 25 hidden units and a 5-layer feedforward network of 50 hidden units per layer each with rectified linear activations. Because we use a binary mask to zero out non-neighboring objects, we implement the encoder layers without bias such that non-neighboring objects do not contribute to the encoder activations. The encoding parameters are shared across all object pairs. The decoder is a five-layer network with 50 hidden units per layer and rectified linear activations after all but the last layer. The NP encoder architecture is the same as the NPE encoder, but without the pairwise layer. The NP decoder architecture is the same as the NPE decoder. The LSTM has three layers of 100 hidden units and a linear layer after the last layer. It has rectified linear activations after each layer. 

We  informally explored several hyperparameters, varying the number of layers from 2 to 5, the hidden dimension from 50 to 100, and learning rates in $\{10^{-5}, 3\times 10^{-5}, 10^{-4}, 3 \times 10^{-4}, 10^{-3}, 3\times 10^{-3}\}$. Though this is far from an exhaustive search, we found that the above hyperparameter settings work well.

\section{Quantitative Analysis}\label{appendix:quant}
\begin{figure}[h!]
  \centering
 \includegraphics[width=\textwidth,height=\textheight,keepaspectratio]{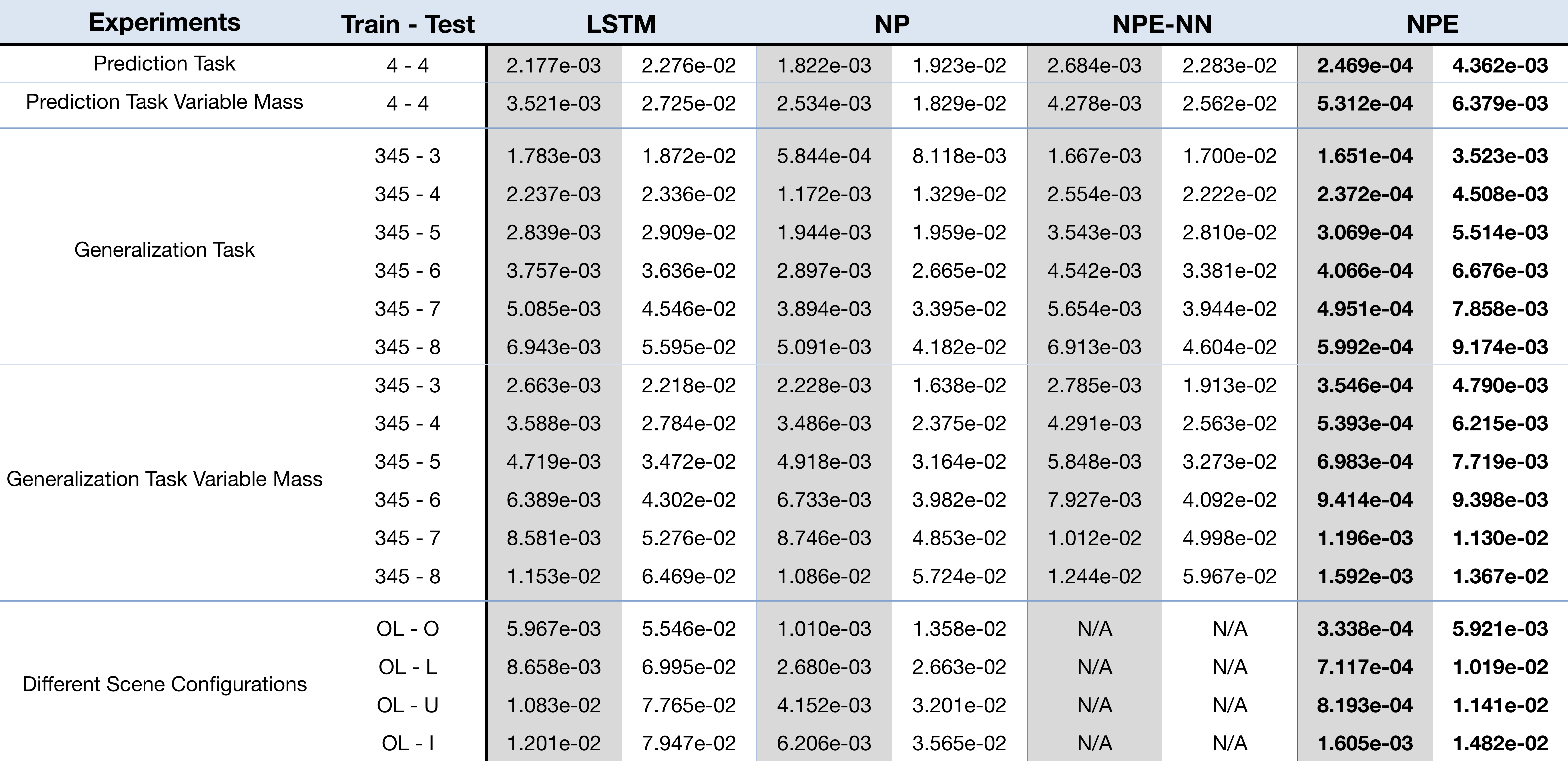}
  \caption{\textbf{Error analysis on velocity and position:} We summarize the error in velocity and position for each train-test variant of each experiment. Normalized velocity MSE is shown in the gray columns (multiplying these values by the maximum velocity of 60 would give the actual velocity in pixels/timestep, where each timestep is about 0.1 seconds). The white columns show the error in  Euclidean distance between the predicted position and the ground truth position of the ball. These have been normalized by the radius of the ball (60 pixels), so multiplying these values by 60 would give the actual Euclidean distance in pixels. The NPE consistently outperforms all baselines by 0.5 to 1 order of magnitude, and this is also reflected in the bottom row of Fig. \ref{figure:ball_quantitative}a,b. Notice that experiments with variable mass exhibit only slightly higher error than their constant-mass variants, even when the variable mass experiments contain masses that differ by a factor of 25. For the experiments with different scene configurations, we do not report error for NPE-NN; the unnecessary computational complexity of operating on over 30 objects, and the degradation in performance without this mask, evident from the other experiments, make the need for the neighborhood mask clear.}
  \label{figure:summary}
\end{figure}

\end{document}